\documentclass[10pt,twocolumn,letterpaper]{article}

\usepackage{cvpr}
\usepackage{times}
\usepackage{epsfig}
\usepackage{graphicx}
\usepackage{subfigure} 
\usepackage{amsmath}
\usepackage{amssymb}
\usepackage{algorithm}
\usepackage{algorithmic}
\usepackage{bm}
\usepackage{multirow}
\usepackage{float}
\usepackage{mathrsfs}
\usepackage{array}
\usepackage{esvect}

\usepackage[pagebackref=true,breaklinks=true,letterpaper=true,colorlinks,bookmarks=false]{hyperref}
\newcommand{\norm}[1]{\left\lVert#1\right\rVert}  
\newcommand{\loss}[1]{\mathcal{L}_\textnormal{#1}}

\newcommand{\nnfn}{g_\theta}
\cvprfinalcopy 
\makeatletter
\renewcommand{\@thesubfigure}{\hskip\subfiglabelskip}
\makeatother

\begin{document}

\title{Domain Alignment with Triplets}

\author{Weijian Deng$^{\dag}$, \ Liang Zheng$^{\ddag}$, \ Jianbin Jiao$^{\dag}\thanks{Corresponding Author}$\\
$^{\dag}$University of Chinese Academy of Sciences \quad $^{\ddag}$ Australian National University\\
\textsl{{\small dengweijian16@mails.ucas.ac.cn}}\\
 }

\maketitle
\begin{abstract}
Deep domain adaptation methods can reduce the distribution discrepancy by learning domain-invariant embedddings. However, these methods only focus on aligning the whole data distributions, without considering the class-level relations among source and target images. Thus, a target embeddings of a bird might be aligned to source embeddings of an airplane. This semantic misalignment can directly degrade the classifier performance on the target dataset.
To alleviate this problem, we present a similarity constrained alignment (SCA) method for unsupervised domain adaptation.
When aligning the distributions in the embedding space, SCA enforces a similarity-preserving constraint to maintain class-level relations among the source and target images, i.e., if a source image and a target image are of the same class label, their corresponding embeddings are supposed to be aligned nearby, and vise versa.
In the absence of target labels, we assign pseudo labels for target images. Given labeled source images and pseudo-labeled target images, the similarity-preserving constraint can be implemented by minimizing the triplet loss.
 With the joint supervision of domain alignment loss and similarity-preserving constraint, we train a network to obtain domain-invariant embeddings with two critical characteristics, intra-class compactness and inter-class separability.
Extensive experiments conducted on the two datasets well demonstrate the effectiveness of SCA.
\end{abstract}
\begin{figure}[t]
  \subfigure{
    \includegraphics[width=0.146\textwidth, height=0.146\textwidth]{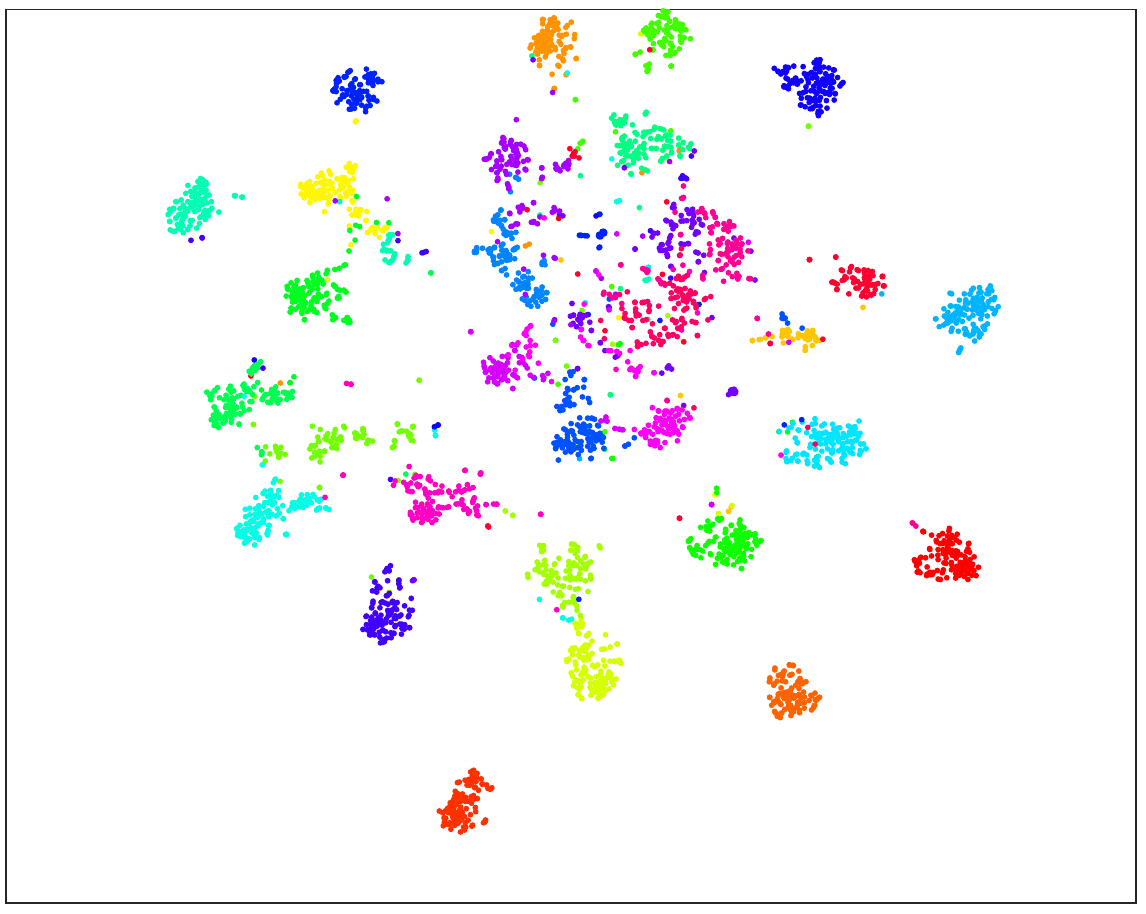}
    \label{fig:tsne1}
  }
  \subfigure{
    \includegraphics[width=0.146\textwidth, height=0.146\textwidth]{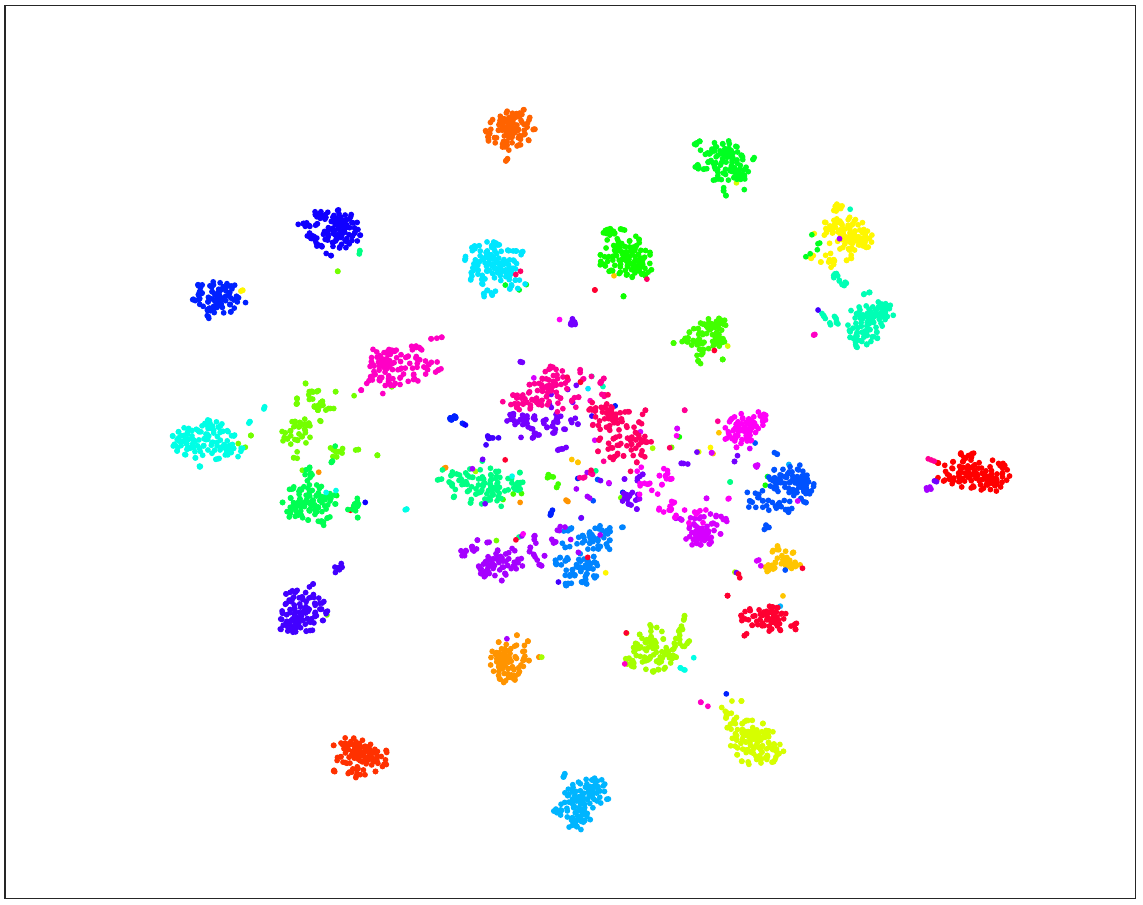}
    \label{fig:tsne2}
  }
  \subfigure{
    \includegraphics[ width=0.146\textwidth, height=0.146\textwidth]{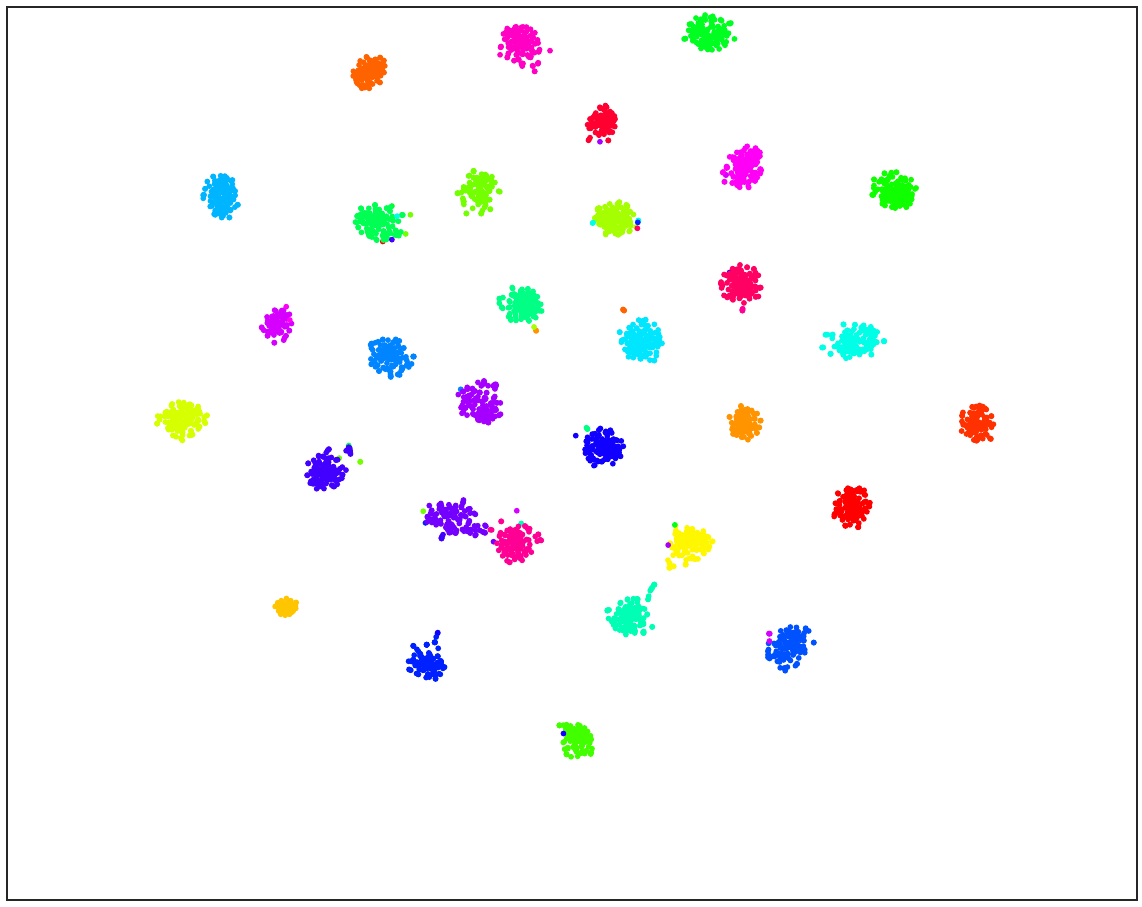}
    \label{fig:tsne3}
  }
  \subfigure[(a) ResNet]{
      \includegraphics[width=0.146\textwidth, height=0.146\textwidth]{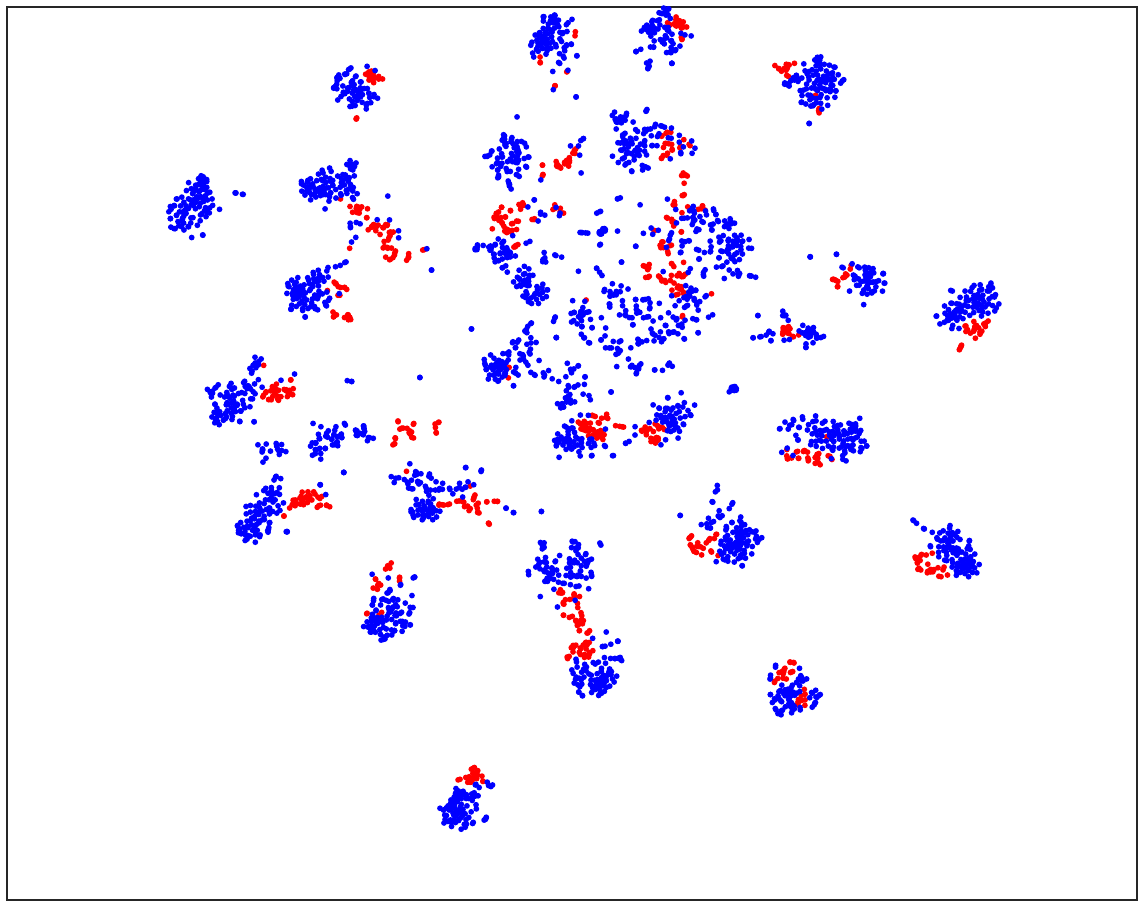}
     \label{fig:tsne} 
  }
  \subfigure[(b) Domain alignment]{
    \includegraphics[width=0.146\textwidth, height=0.146\textwidth]{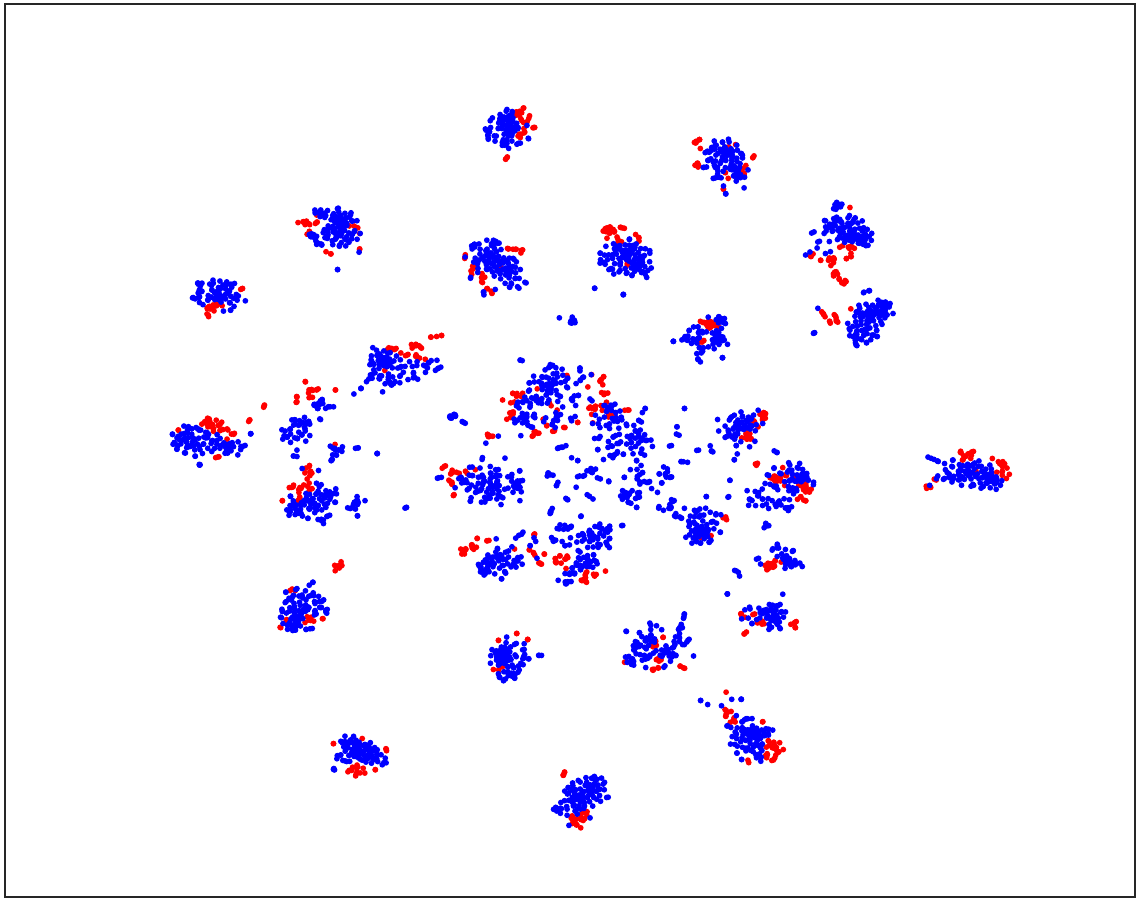}
  }
  \subfigure[(c) SCA]{
    \includegraphics[width=0.146\textwidth, height=0.146\textwidth]{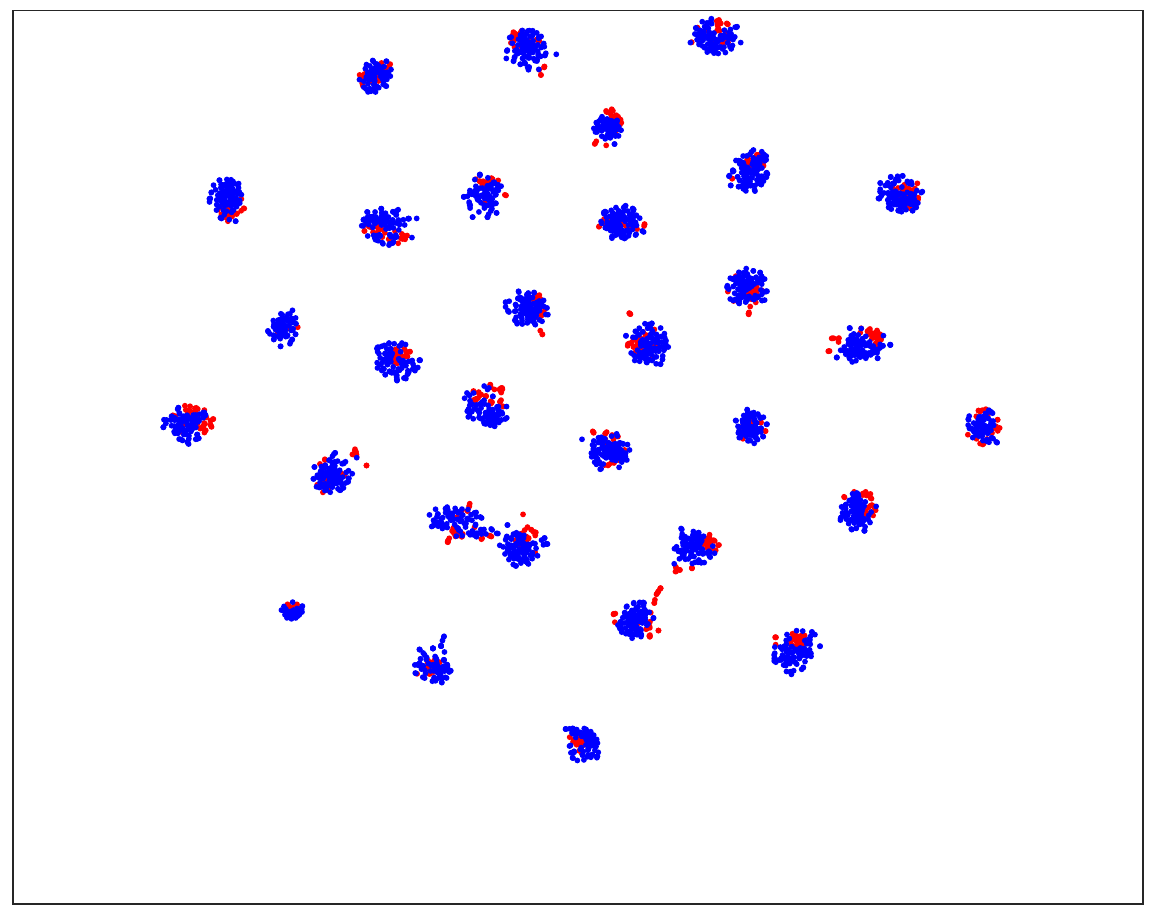}
  }
  \caption{Visualization of cross-domain embeddings for task \textbf{A} $\rightarrow$ \textbf{W} on Office-31 \cite{SaenkoKFD10}. We present the 2D visualization of t-SNE for embeddings learned by (a) ResNet (trained on source images only), (b) domain alignment (based on JMMD \cite{JAN}), and (c) SCA (ours). For the \textbf{first row}, different colors denote data of different object categories. For the \textbf{second row}, red color represents the data of \textbf{W}, and blue color represents data of \textbf{A}. Under SCA, different classes are well-separated, and the two domains are well-aligned on the class level. Best viewed in color.}
\end{figure}
\section{Introduction} \label{Introduction}
In many real-world application of visual recognition, the training and testing data distributions are often different due to \emph{dataset bias} \cite{DBLP:conf/cvpr/TorralbaE11}. This distribution discrepancy decreases the generalization capability of the learned visual representations. One example is that the model trained on synthetic images fails to generalize well on the real-world images. To eliminate the effect of the dataset bias, a common used strategy is unsupervised domain adaption (UDA). 
 In UDA, we are provided with a labeled source dataset and an unlabeled target dataset, and the goal is to learn a model on the source dataset which minimizes the test error on the target dataset.

In literature, recent UDA methods \cite{GaninUAGLLML16, JAN, GaninL15, TzengHSD17, TzengHZSD14, DAN} adopt deep neural networks to learn a shared embedding space where the distribution discrepancy can be reduced. These methods typically involve two objectives: 1) learn embeddings that maintain a low classification error on the source dataset; 2) make embeddings domain-invariant, such that the classifier trained on the source can be directly used on the target dataset. To learn domain-invariant embeddings, recent methods usually minimize some measure of domain variance \cite{TzengHZSD14, JAN, DAN} (such as correlation distance \cite{SunS16}) or adopt the adversarial learning \cite{GaninUAGLLML16, GaninL15, TzengHSD17}. 
However, this line of methods have an intrinsic limitation: they only focus on reducing the global distribution discrepancy, without exploiting the class-level relations among the source and target images. 
Thus, even with perfect distribution alignment, the images with different labels from different domains might be misaligned nearby in the embedding space. 
As shown in Fig. \ref{fig:tsne2}, domain-level alignment (based on JMMD \cite{JAN}) has the ability to reduce distribution discrepancy. However, these exists the semantic misalignment problem in the aligned embeddings. For examples, some samples from different classes are mapped nearby in the embedding space. This semantic misalignment is detrimental to the classifier performance on the target dataset. 

Motivated by this problem, we present a similarity constrained alignment (SCA) method for UDA. The working mechanism of SCA is that it can align the distributions, while preserving the class-level relations among source and target images. Specifically, we add a similarity-preserving constraint for the source and target images during domain alignment. The impact of the similarity-preserving constraint is two-fold. 1) \emph{Class unification}: images with same labels should be pulled together in the embedding space; 2)\emph{ class separation}: images with different labels should be pushed apart.  In practice, the similarity-preserving constraint can be implemented by minimizing the triplet loss \cite{SchroffKP15}.
During training, SCA learns domain-invariant embeddings by optimizing  an objective that includes both the domain confusion loss and the triplet loss \cite{SchroffKP15}.
First, the domain confusion loss aims at mapping the source and target distributions into a shared feature space. Several existing methods can be directly used to achieve this goal. In this paper, we adopt JMMD \cite{JAN} to align the data distributions.
Second, the triplet loss is to enhance the discriminative ability of the deeply learned embeddings, so that source and target embeddings possess the properties of intra-class compactness and inter-class separability. 

Unfortunately, the target dataset is totally unlabeled, so the similarity-preserving constraint cannot be directly imposed for the source and target images.
In the absence of target labels, we use a classifier trained on source images to assign pseudo labels for target images. To eliminate the influence of the incorrectly assigned images, we only select images with high predicted scores for training.
Given labeled source images and pseudo-labeled target images, we utilize the triplet loss \cite{SchroffKP15} to constrain their similarity in the embedding space. Specifically, if a source image and a target image are with the same class label, their corresponding embeddings are supposed to be aligned nearby, and vise versa.
In this manner, the semantic misalignment problem can be alleviated. 
As shown in Fig. \ref{fig:tsne3}, we observe that the embeddings learned by our method preserve the two class-level relations: 1) the embeddings that belong to the same class are close (class unification); 2) the embeddings that belong to different classes are separated well (class separation).
Based on the domain-invariant embedddings learned by SCA, the classifier can generalize well on the target dataset.

To summarize, this paper is featured in three aspects. First, to our knowledge, this is an early work that explores the class-level relations across domains under the UDA setting. Second, by consolidating the idea of domain-level alignment and metric learning, this paper presents a novel similarity constrained alignment (SCA) method for UDA. SCA attempts to reduce the distribution discrepancy while preserving the underlying difference and commonness among source and target images. Thus, the class-level misalignment problem can be alleviated. Third, extensive experiment results demonstrate that the proposed method improves the generalization ability of the learned classifier. Moreover, the proposed method is capable of producing competitive accuracy to state-of-the-art methods on two UDA benchmarks.
\begin{figure*}[!t]
\setlength{\abovecaptionskip}{-0.2cm} 
\begin{center}
\includegraphics[width=0.88\linewidth]{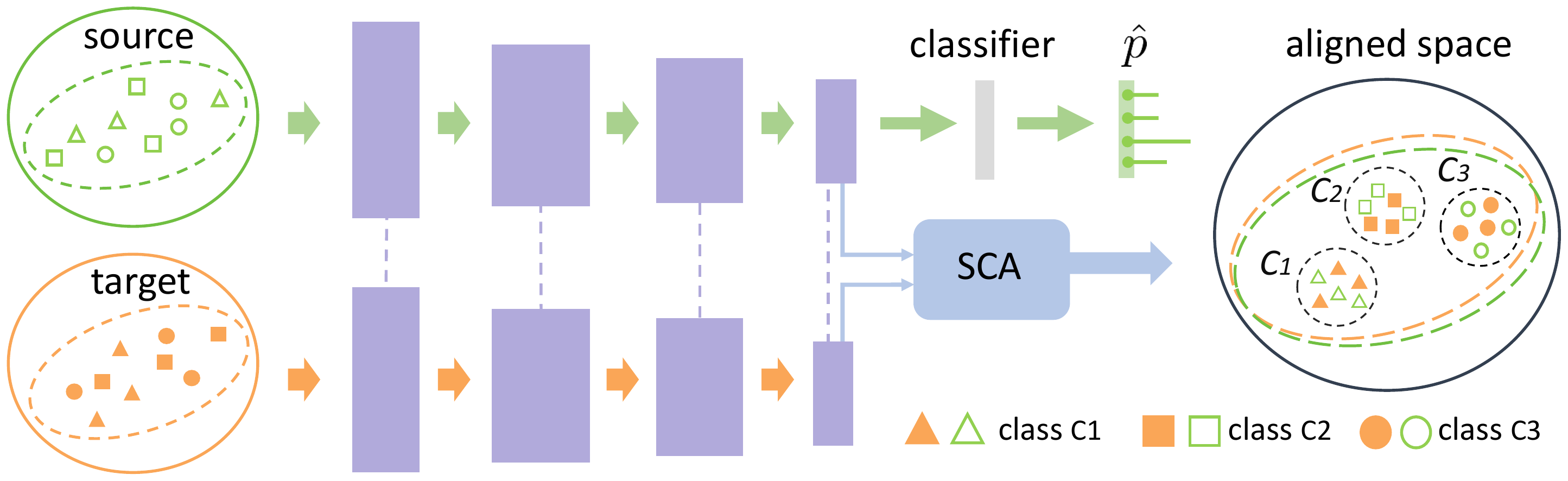}
\end{center}
\caption{Framework of the similarity constrained alignment (SCA) method. SCA has the ability to align the distribution, while preserving the class-level relations among source and target images. Thus,  if a source image and a target image are with the same class label, their corresponding embeddings are supposed to be aligned nearby, and vise versa.
Due to the target dataset is unlabeled, we assign pseudo labels for the target images (see Section \ref{la}). In this figure, different colors denote different domain distributions and different shapes represent different classes. }
\label{fig:motivation}
\end{figure*}
\section{Related Work}
Many methods are proposed to solve the domain adaptation problem. This section briefly reviews works that are closely related to our paper.

\textbf{Unsupervised domain adaptation.} Unsupervised domain adaptation methods attempt to minimize the shift between source and target data distributions. Some methods focus on learning a mapping function between source and target distributions \cite{KulisSD11, GopalanLC11, FernandoHST13, DBLP:conf/aaai/SunFS16}. In \cite{DBLP:conf/aaai/SunFS16}, Correlation Alignment is proposed to match the two distributions. In \cite{FernandoHST13}. The source and target domain are aligned in the subspace described by Eigenvectors.

Other methods seek to find a shared feature space for source and target distributions \cite{GaninL15, TzengHZSD14, DAN, JAN}. Long \etal \cite{DAN} and Tzeng \etal \cite{TzengHZSD14} utilize the maximum mean discrepancy (MMD) metric \cite{GrettonBRSS06}  to learn a share feature representations.  Moreover, the joint maximum mean discrepancy (JMMD) \cite{JAN} is proposed to align the joint distributions of multi-layers across domains. Recent methods \cite{GaninUAGLLML16, GaninL15, TzengHSD17, zhang2018collaborative, CaoMLW18, PeiCLW18} adopt adversarial learning \cite{GoodfellowPMXWOCB14} to learn representations that are not able to distinguish between domains. 
The gradient reversal algorithm (RevGrad) \cite{GaninL15} is proposed to learn the domain invariant feature.
Tzeng \etal \cite{TzengHSD17} propose a generalized framework for adversarial domain adaptation.
Pei \etal \cite{PeiCLW18} propose a multi-domain adversarial network for fine-grained distribution alignment. SimNet \cite{simnet} proposes to classify an image by computing its similarity to prototype representations of each category.
Some methods \cite{ hoffman2018cycada, DBLP:journals/corr/BousmalisSDEK16, DBLP:conf/nips/LiuT16, image-image18} use the adversarial learning to learn a transformation in the pixel space from one domain to another. CYCADA \cite{ hoffman2018cycada} maps samples across domains at both pixel level and feature level.
In this paper, we also attempt to reduce the distribution discrepancy, and we are more concerned with preserving the class-level relations among the source and target datasets.

\textbf{Self-training.} Our method is related to self-training, a strategy in which the predictions of a classifier on the unlabeled data are used to retrain the classifier \cite{LiF10, ChenSG13, LaineA16, zhang2018collaborative, distill, kang2018deep}. The assumption of self-training is that an image with the high predicted score is more likely to be classified correctly. In unsupervised domain adaptation, some methods \cite{zhang2018collaborative, ChenWB11, SaitoUH17} use pseudo-labeled images to improve classifier accuracy on the target dataset. Zhang \etal \cite{zhang2018collaborative} propose a progressive way to select pseudo-labeled images for training the classifier. Chen \etal \cite{ChenWB11} use two classifiers to assign labels for target images. Saito \cite{SaitoUH17} adopt three asymmetric classifiers to improve the quality of pseudo labels.
Unlike these methods, we leverage the selected images with their pseudo-labels for semantic alignment instead of retraining the classifier. This practice provides a new way to utilize unlabeled data for learning feature representations.

\textbf{Deep Metric learning.} Deep metric learning \cite{ChopraHL05, GoldbergerRHS04, WeinbergerS09, SongXJS16, SchroffKP15, HuLT14} aims to learn discriminative embeddings such that similar samples are nearer and different samples are further apart from each other. The most widely used loss functions for deep metric learning are the contrastive Loss \cite{ChopraHL05} and triplet loss \cite{SchroffKP15}. 
The problem settings of these works are different from ours. We aim to reduce the distribution discrepancy and utilize the triplet loss \cite{SchroffKP15} to preserve the class-level relations among images from the two domains. Since the target domain is unlabeled, we assign pseudo labels for the target images.

\section{Proposed Method}
\subsection{Overview}
In UDA, we are provided with a set of labeled images from the source dataset and a set of unlabeled images from the target dataset, where the data distributions of the two datasets are different.  For the source dataset, we denote it as $\mathcal{D}_s = \{(\mathbf{x}_i^s,{\bf y}^s_i)\}_{i=1}^{n_s}$, where $\mathbf{x}_i^s$ is the $i$-th source image, ${\bf y}^s_i$ is its label, and $n_{s}$ is the total number of images on the source dataset. Similarly, we denote the target dataset as ${{\cal D}_t} = \{ {\bf{x}}_j^t\} _{j = 1}^{{n_t}}$, where $\mathbf{x}_i^t$ is the $i$-th target image and $n_{t}$ is the total number of images on the target dataset. 
Our goal is to leverage labeled source images and unlabeled target images to learn a classifier that can generalize well on the target dataset.

It is worth repeating that, for UDA, we not only deal with the whole distribution discrepancy that caused by the dataset bias. We also consider preserving the class-level relations among source and target images. 
To this end, we present a similarity constrained alignment (SCA) method for UDA. As shown in Fig. \ref{fig:motivation}, the goal of SCA is two-fold, 1) it learns a domain-invariant embedding space to align the whole distributions; 2) it preserve the underlying difference and commonness among source and target images. Based on the learned embeddings, we can train a classifier that can generalize well on the target dataset.

In Section \ref{ga}, we briefly describe the domain-level alignment method used in this paper. In Section \ref{la}, we introduce the similarity-preserving scheme. In Section \ref{Discussion}, we have a discussion about the proposed method.

\subsection{Similarity Preserving Alignment}
In this paper, we utilize the deep convolution neural network to learn the classifier. For K-way classification with a cross-entropy loss, this is corresponding to,
\begin{equation}\label{loss:cross-entropy}
 {\mathcal{L}_{c}} = \frac{1}{{{n_s}}}\sum\limits_{i = 1}^{{n_s}} {L\left( {f\left( {{\nnfn(\mathbf{x}}_i^s)} \right),{\mathbf{y}}_i^s} \right)},
\end{equation}
where $L(\cdot, \cdot)$ is the cross-entropy loss, where  $\nnfn(\cdot)$ is the feature extractor, and $f(\cdot)$ is the classifier trained on the source dataset.

In general, the classifier $f(\cdot)$ is a simple fully-connected network followed by a softmax over the classes .
Due the dataset bias \cite{DBLP:conf/cvpr/TorralbaE11}, the classifier trained on the source dataset often fails to generalize well on the target dataset. To alleviate this problem, we present a similarity constrained alignment (SCA) method. SCA can eliminate the distribution discrepancy, while preserving the underlying difference and commonness among source and target images. In practice, SCA learn domain-invariant embeddings by optimizing over an objective that includes both the domain-level alignment loss and the similarity-preserving loss. 

\subsubsection{Domain-level alignment} \label{ga}
Domain-level alignment focuses on reducing the whole distribution discrepancy between the source and target datasets. In the community, recent deep domain adaptation methods utilize a domain confusion loss to align the distributions. These methods usually adopt the discrepancy-based metric \cite{GrettonBRSS06, JAN} or adversarial adaptation \cite{GaninL15} to design the domain confusion loss function.

 Following the practice in \cite{JAN}, we build the domain-level alignment loss by using the JMMD metric. 
The JMMD formally reduces the discrepancy in the joint distributions of the activations in domain-specific layers $\mathcal{L}$, i.e. $P(\mathbf{Z}^{s1},\ldots,\mathbf{Z}^{s|\mathcal{L}|})$ and $Q(\mathbf{Z}^{t1},\ldots,\mathbf{Z}^{t|\mathcal{L}|})$. Thus, the loss function of domain-level alignment is written as, 
\begin{equation}
\small
\begin{aligned}
  {\mathcal{L}_{d}}& = {\frac{2}{n}\sum\limits_{i = 1}^{n/2} {\left( {\prod\limits_{\ell  \in \mathcal{L}} {{k^\ell }( {{\mathbf{z}}_{2i - 1}^{s\ell },{\mathbf{z}}_{2i}^{s\ell }} )}  + \prod\limits_{\ell  \in \mathcal{L}} {{k^\ell }( {{\mathbf{z}}_{2i - 1}^{t\ell },{\mathbf{z}}_{2i}^{t\ell }} )} } \right)}}  \\
   & - {\frac{2}{n}\sum\limits_{i = 1}^{n/2} {\left( {\prod\limits_{\ell  \in \mathcal{L}} {{k^\ell }( {{\mathbf{z}}_{2i - 1}^{s\ell },{\mathbf{z}}_{2i}^{t\ell }} )}  + \prod\limits_{\ell  \in \mathcal{L}} {{k^\ell }( {{\mathbf{z}}_{2i - 1}^{t\ell },{\mathbf{z}}_{2i}^{s\ell }} )} } \right)}}, \\ 
\end{aligned}\label{loss:dataset-level}
\end{equation}
where $n=n_s$, ${\mathbf{z}}^{t\ell }$ denotes the activations of the target image in the layer ${\ell }$, and ${\mathbf{z}}^{s\ell }$ denotes the activations of the source image in the layer ${\ell }$. $k^\ell$ is the kernel function in a reproducing kernel Hilbert space (RKHS).

We adopt the ResNet-50 \cite{resnet} as the backbone network.  We discard its last layer and add two fully connected layers (a bottleneck layer, and a classifier layer) for our task. In practice, we align the joint distributions of the activations in two newly added layers.

\subsubsection{Similarity-constrained Scheme} \label{la}
Domain-level alignment only aims at reducing the whole distribution discrepancy, but it can mix up the class-level relations among the source and target images.  Consequently,  there exists a semantic misalignment problem, \ie, source images of class A might be falsely aligned to target images of class B in the embedding space. This semantic misalignment problem directly degrades accuracy on the target dataset. 
To mitigate this problem, we should consider the class-level relations of images across two datasets. In this paper, we propose to preserve the underlying difference and commonness among images during the domain alignment.

\textbf{Class-level relations.} 
A general assumption behind the similarity-preserving alignment is that if a source image and a target image are with the same class label, their corresponding embeddings are supposed to be aligned nearby, and vise versa.
On the top of domain-level alignment, we add a similarity-preserving constraint to maintain two class-level relations among source and target images.
In this paper, the two class-level relations are defined as follow.
\begin{itemize}
\setlength{\itemsep}{-0pt}
\item \emph{Class separation.} Images from different domains and with different labels, should be mapped  far apart in the embedding space.
\item \emph{Class unification.} Images from different domains but with same labels, should be mapped nearby in the embedding space
\end{itemize}

\textbf{Similarity-preserving loss function.}
To mitigate the semantic misalignment problem, we want images to preserve the above class-level relations during the domain-level alignment. 
Let $D_{i,j} = \norm{\nnfn(x_i) - \nnfn(x_j)}_2^2$ measures the distance between two images in the feature space, where  $\nnfn(\cdot)$ is the feature extractor. If $x_i$ and $x_j$ are with the same label, we want $D_{i,j}$ to be small, corresponding to the class unification. If $x_i$ and $x_j$ are with different labels, we want $D_{i,j}$ to be large, corresponding to the class separation. 

Based on the above analysis, we utilize the triplet loss \cite{SchroffKP15} to achieve similarity-preserving constraint. 
 Given an \emph{anchor} image $x_a$, a \emph{positive} image $x_p$, and a \emph{negative} image $x_p$, we minimize the loss,
    \begin{equation}\label{eq:loss_trip}
        \loss{s}(\theta) = \sum\limits_{\substack{a,p,n \\ y_a = y_p \neq y_n}} \left[m + D_{a,p} - D_{a,n}\right]_+,
    \end{equation}
where $x_a$ and $x_p$ is a positive pair (their labels $y_a$ and $y_p$ are same), $x_a$ and $x_p$ is a negative pair (their labels $y_a$ and $y_n$ are different). $m$ is the margin that is enforced between positive and negative pairs.

This loss encourages the distance between $x_a$ and positive image $x_p$ to be smaller than the distance between $x_a$ and negative $x_n$ by the enforced margin $m$.

\textbf{Training data construction.}
The similarity-preserving loss supervises the embedding learning, so that class-level relations among source and target images can be preserved.
When optimizing the similarity-preserving loss, we should pay attention to two crucial things, 1) the target dataset is totally unlabeled; 2) the construction of training triplet samples is non-trivial.
For these two things, we propose corresponding techniques.

\textbf{(i) Label estimation for unlabeled target data.}
The target dataset is totally unlabeled, so the semantic relations cannot be directly built. In the absence of target labels, we use a classifer pre-trained on the source images to assign labels for unlabeled target images. 

To ensure the accuracy of the pseudo label, we adopt three tactics.
(a) \emph{domain-level alignment.} When pre-training the classifier, we also utilize the dataset-level to reduce the harmful influence of dataset bias. This practice improves the performance of the classifier on the target dataset, so that more accurate pseudo labels can be gained.
(b) \emph{Threshold $T$}. Intuitively, the image with the high predicted score is more likely to be classified correctly.
Thus, we only select target images with predicted scores above a high threshold $T$ for building the semantic relations. Note that the threshold $T$ is constant during training.
(c) \emph{Progressive selection.} With the help of the similarity-preserving alignment, the classifier will improve itself during training. This motivates us to re-assign the label for the target image every several iterations (K). By doing so, the target images can be progressively selected for the class-level alignment.

\textbf{(ii) Sample triplet images.}  Given labeled source images and pseudo-labeled target images, we now introduce the way to construct triplet samples. The possible number of triplets is large, and optimizing all triplets is computationally infeasible. 
To avoid this problem, we follow the sampling strategy in \cite{HermansBL17}. For the labeled source images, we randomly select $C$ classes and randomly select $K$ images of each class. In this way, we select $CK$ source images. Similarly, we select $CK$ pseudo-labeled target images. Thus, we get a mini-batch of 2$CK$ training images and perform triplet sampling in each mini-batch.

\subsubsection{Overall objective}\label{all}
We present a similarity constrained alignment (SCA) for UDA. During the training, SCA jointly optimizes an objective that includes both a domain-level alignment loss and a similarity-preserving loss, such that more discriminative domain-invariant embeddings can be gained. On the top of learned embeddings, we can train a classifier that generalizes well on the target dataset.
The final objective of SCA is written as, 
\begin{equation}\label{loss:all}
 {\mathcal{L}_{can}} =  {\mathcal{L}_{c}} + \alpha {\mathcal{L}_{d}} + \beta {\mathcal{L}_{s}},
\end{equation}
where $ {\mathcal{L}_{c}}$ is the classification loss, ${\mathcal{L}_{d}} $ is the domain-level domain alignment, and ${\mathcal{L}_{s}}$ is the similarity-preserving loss. The $\alpha$ and the $ \beta$ control the relative importance of domain-level alignment and similarity preservation, respectively.

\subsection{Discussion} \label{Discussion}
\textbf{Collaborative working mechanism.} The working mechanism of SCA is that it can align the distributions, while preserving the class-level relations among source and target images.  On the one hand, if we only use the domain-level alignment to reduce the distribution discrepancy, the resulting embeddings would exist the semantic misalignment problem. On the other hand, the similarity-preserving constraint can map a source image and a target image nearby, if they are with the same class label. Thus, the similarity-preserving constraint can be viewed as the class-level distribution alignment.
 With the collaborative supervision of them, we can reduce the distribution at both domain level and class level, \ie, learning domain-invariant embedddings that preserve the class-level relations. In our experiment, we validate this collaborative working mechanism. Moreover, we also study the impact of only adopting the similarity-preserving constraint on the transfer accuracy.

Closely related to our work, Motiian \etal \cite{MotiianPAD17} also study the class-level alignment. Our work is different from \cite{MotiianPAD17} in two aspects, 1) the setting of \cite{MotiianPAD17} is supervised domain adaptation, where the labeled target images are available; 2) the authors of \cite{MotiianPAD17} do not consider the domain-level alignment, while our work collaboratively aligns the distributions at both domain and class level.

\textbf{Label estimation.} To construct class-level relations among the source and target images, we need to estimate the labels of unlabeled target images. In this paper, we simply adopt a classifier pre-trained on the source images to assign pseudo labels for unlabeled target images. 
We only select target images with their scores above a certain threshold $T$. Note that we do not adaptively adjust the threshold $T$ as in \cite{zhang2018collaborative}.  In practice, we set the threshold $T$ a high value  (0.9) to guarantee that the selected samples are more likely to be predicted correctly. 

During the training, the classifier will gradually improve itself, so we re-assign pseudo labels every several iterations. In this way, more and more target images will be progressively selected for training.

\textbf{How to use pseudo-labeled target images?} Existing methods \cite{LiF10, ChenSG13, LaineA16, zhang2018collaborative, distill, kang2018deep} usually utilize the pseudo-labeled target images for training classifier directly.
In this paper, the pseudo-labeled images are \textbf{not used} for training the classier, but for building the class-level relations.  We argue that there exist a set of wrongly pseudo-labeled images, which can directly bring a bad influence to the classifier. 
To avoid this problem, we use selected target images for optimizing the similarity-preserving loss function.

Moreover, as analyzed in \cite{LiuWYLRS17, WenZL016}, cross-entropy loss encourage the features of different classes staying apart. Thus, using selected target images for training classifier can be viewed as an indirect way to preserve the class separation relation. However, the cross-entropy loss does not consider the class unification relation.
In contrast, we adopt pseudo-labeled target images and source images for constructing both class unification and separation relations.
\section{Experimental Evaluation}

\subsection{Datasets}
We evaluate the proposed unsupervised domain adaptation method on two datasets, \ie, Office-31 \cite{SaenkoKFD10} and ImageCLEF-DA\footnote{\url{http://imageclef.org/2014/adaptation}}.

\textbf{Office-31}  is a widely used benchmark for visual domain adaptation. It contains 4,652 images and 31 categories collected from three distinct domains: \textit{Amazon} (\textbf{A}), \textit{Webcam} (\textbf{W}) and \textit{DSLR} (\textbf{D}). The images in DSLR are captured with a digital SLR camera and have high resolution. Amazon consists of images downloaded from online merchants (www.amazon.com). These images are of products at medium resolution. The images in Webcam are collected by a web camera, and they are of low resolution.
We evaluate the proposed method across six transfer tasks \textbf{A} $\rightarrow$ \textbf{W}, \textbf{D} $\rightarrow$ \textbf{W},  \textbf{W} $\rightarrow$ \textbf{D}, \textbf{A} $\rightarrow$ \textbf{D}, \textbf{D} $\rightarrow$ \textbf{A} and \textbf{W} $\rightarrow$ \textbf{A}. We report the results following the protocol in \cite{DAN}.

\begin{figure}[t]
\setlength{\abovecaptionskip}{-0.1cm} 
\setlength{\belowcaptionskip}{-0.2cm}
\begin{center}
\includegraphics[width=1 \linewidth]{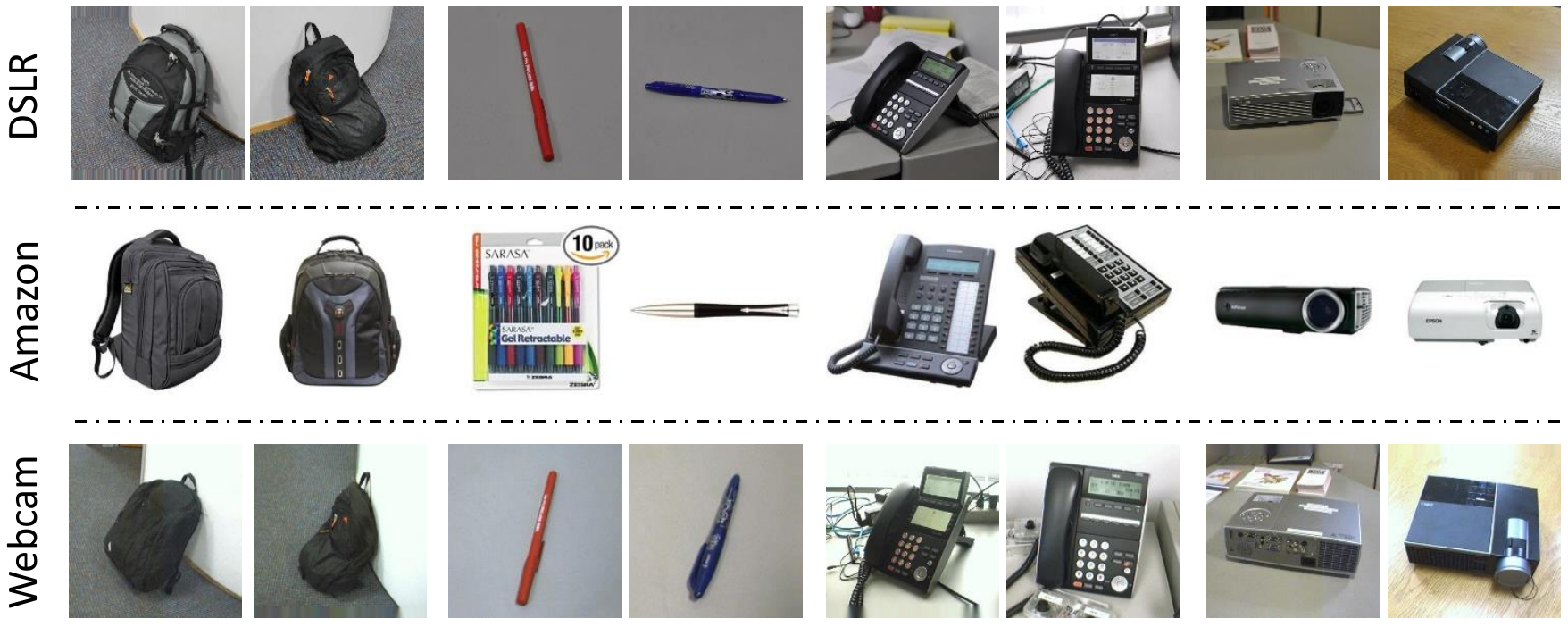}
\end{center}
\caption{Visual examples of the Office-31 dataset. From top to bottom: DSLR images (high-resolution), Amazon images (medium-resolution), and Webcam images (low-resolution).}
\label{fig:office}
\end{figure}

\begin{figure}[t]
\setlength{\abovecaptionskip}{-0.1cm} 
\setlength{\belowcaptionskip}{-0.2cm}
\begin{center}
\includegraphics[width=1 \linewidth]{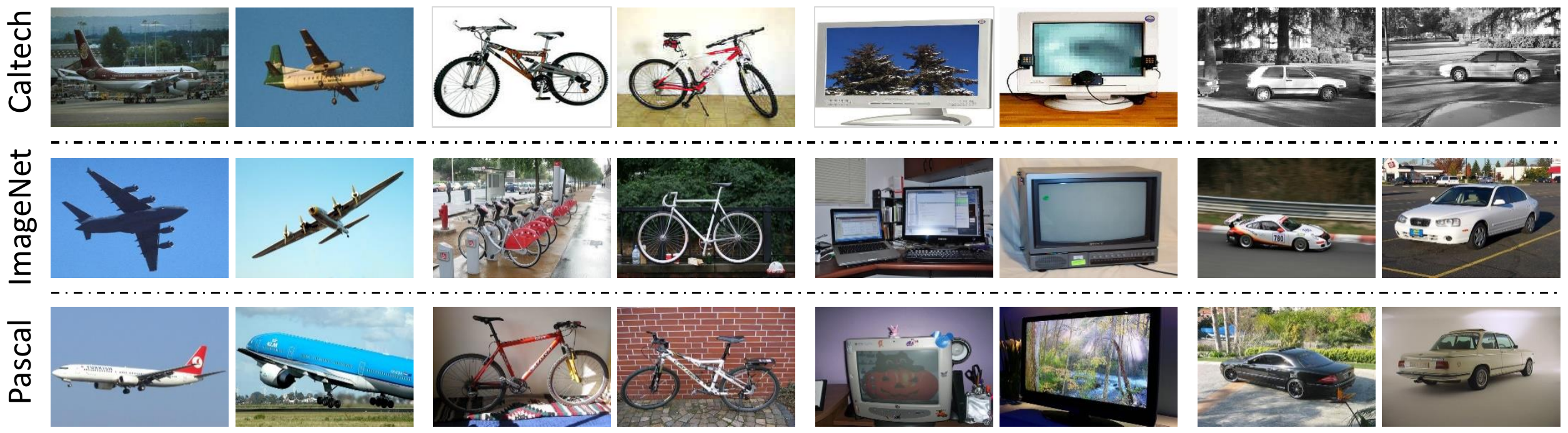}
\end{center}
\caption{Visual examples of the ImageCLEF-DA dataset. From top to bottom: Caltech-256 images, ImageNet ILSVRC 2012 images, and Pascal VOC 2012 images.}
\label{fig:image}
\end{figure}

\textbf{ImageCLEF-DA} is a benchmark dataset for ImageCLEF 2014 domain adaptation challenge. It contains three subsets, including \textit{Caltech-256} (\textbf{C}), \textit{ImageNet ILSVRC 2012} (\textbf{I}), and \textit{Pascal VOC 2012} (\textbf{P}), and each subset is considered as a domain. There are 12 categories and each categories contains 50 images. We use all domain combinations and build 6 transfer tasks: \textbf{I} $\rightarrow$ \textbf{P}, \textbf{P} $\rightarrow$ \textbf{I}, \textbf{I} $\rightarrow$ \textbf{C}, \textbf{C} $\rightarrow$ \textbf{I}, \textbf{C} $\rightarrow$ \textbf{P}, and \textbf{P} $\rightarrow$ \textbf{C}. We report the results following the protocol in \cite{JAN}.
Sample images of the Office-31 and ImageCLEF-DA are shown in Fig \ref{fig:office} and Fig \ref{fig:image}, respectively. Through these images, we can observe the dataset bias discussed in \cite{DBLP:conf/cvpr/TorralbaE11}.

\begin{algorithm}[t]
  \caption{Similarity Constrained Alignment (SCA).}
  \begin{algorithmic}[1]\label{can}
  \INPUTS
    \STATE source images and labels  $\{(\mathbf{x}_i^s,{\bf y}^s_i)\}_{i=1}^{n_s}$, \\
    unlabeled target images $\{ {\bf{x}}_j^t\} _{j = 1}^{{n_t}}$.\\
   threshold ($T$), max number of steps ($S$), and number of SCA updates per step ($K$).
  \ENDINPUTS
  \textbf{stage 1: pre-train a classifier}:
   \STATE train a classifer by minimizing Eq. \ref{loss:cross-entropy} and Eq. \ref{loss:dataset-level}.\\
  \textbf{stage 2: class-level alignment}:
    \FOR{s=1; s $\le$ $S$; s $++$}
      \STATE use classifer to assign pseudo labels for target images with predicted score above $T$.
      \FOR{k =1; k $\le$ $K$; k $++$}
      \STATE train SCA by minimizing Eq. \ref{loss:all}.
     \ENDFOR 
    \ENDFOR
  \end{algorithmic}
\end{algorithm}

\subsection{Implementation Details}
We implement our method on pytorch framework, and fine-tune from ResNet-50 model \cite{resnet} pre-trained on the ILSVRC 2012 dataset \cite{ILSVRC15}. All the images are resized to $256 \times 128$. We discard its last layer and add two fully connected layers for our task. The first layer has 256 units, and the second goes down to the number of training classes.
During training, we adopt random flipping and random cropping as data augmentation methods.
We use stochastic gradient descent (SGD) for optimization, and adopt the same INV learning rate strategy as in RevGrad \cite{GaninL15}. The learning rate decreases gradually after each iteration from 0.001, the momentum is set to 0.9, and the weight decay is set to 0.0004. We set $\alpha =1$ and $\beta =1$  in Eq. \ref{loss:all}.

We adopt a two-stage training procedure: we first initial the classifier by minimizing Eq. \ref{loss:cross-entropy} and Eq. \ref{loss:dataset-level}, then train the whole network by minimizing Eq. \ref{loss:all}. The training procedure is summarized in Algorithm \ref{can}. For the stage one training, we train the network for 5000 iterations. 
For the stage two training, we training the remaining 30000 iterations. We set threshold  $T = 0.9$, max number of step $S = 15$, and number of SCA updates per step $K =2000$ .

\subsection{Experimental Results}

\textbf{Compared Apporaches.} In this section, we mainly compare the proposed method with several state-of-the-art methods, including DAN \cite{DAN}, RTN \cite{RTN}, JAN \cite{JAN}, RevGrad \cite{GaninL15}, MADA \cite{PeiCLW18},  SimNet \cite {simnet},  iCAN \cite{zhang2018collaborative}, and CDAN \cite{long2017conditional}. These methods are all based on the deep neural network (ResNet-50 \cite{resnet}) to learn domain-invariant embeddings. For the fair comparison, the results of these methods are directly reported from their original papers.

\setlength{\tabcolsep}{2.2pt}
\begin{table}[!t]
  \centering
  \label{table:office31}
  \begin{footnotesize}
  \begin{tabular}{l| c|c| c| c| c| c|c}
    \hline
   Method & A$\rightarrow$W & A$\rightarrow$D & W$\rightarrow$A & W$\rightarrow$D & D$\rightarrow$A & D$\rightarrow$W & Avg. \\
    \hline
    \hline
    ResNet-50 \cite{resnet}&72.5&73.6&59.9&99.3&61.0&93.6&76.7 \\
    DAN \cite{DAN}&80.5&78.6&62.8&99.6&63.6&97.1&80.4\\
    RTN \cite{RTN}&84.5&77.5&64.8&99.4&66.2&96.8&81.6\\
    JAN \cite{JAN}&85.8&85.0&70.0&99.7&68.9&96.7&84.4\\
    RevGrad \cite{GaninL15}&82.0&79.7&67.4&99.1&68.2&96.9&82.2\\
    MADA \cite{PeiCLW18}&90.0&87.8&66.4&99.6&70.3&97.4&85.2\\
    SimNet \cite {simnet}&88.6&85.3&71.8&99.7& \textbf{73.4}&98.2&86.2\\
    iCAN \cite{zhang2018collaborative}&92.5&90.1&69.9& \textbf{100.0}&72.1& \textbf{98.8}&87.2\\
    CDAN-RM \cite{long2017conditional}&93.0&89.2&69.4& \textbf{100.0}&70.2&98.4&86.7\\
    CDAN-M \cite{long2017conditional}&93.1& \textbf{93.4}&70.3& \textbf{100.0}&71.0&98.6& \textbf{87.7}\\
    \hline
   SCA & \textbf{93.5} &89.5& \textbf{72.7}& \textbf{100.0}&72.4&97.5&87.6\\
    \hline
  \end{tabular}
  \end{footnotesize}
  \caption{Comparison of different methods for unsupervised domain adaptation on the Office-31 dataset. The best results are in \textbf{bold}.}
    \label{office}
\end{table}

\setlength{\tabcolsep}{2.8pt}
\begin{table}[!t]
\begin{center}
  \label{table:office31}
  \begin{small}
  \begin{tabular}{l| c| c| c| c| c| c|c}
    \hline
    Method & I$\rightarrow$P & P$\rightarrow$I & I$\rightarrow$C & C$\rightarrow$I & C$\rightarrow$P & P$\rightarrow$C & Avg. \\
    \hline
    \hline
    ResNet-50 \cite{resnet}&74.8&82.9&91.5&78.0&66.2&87.2&80.1\\
    DAN \cite{DAN} &74.5&82.2&92.8&86.3&69.2&89.8&82.5\\
    RTN \cite{RTN}&74.6&85.8&94.3&85.9&71.7&91.2&83.9\\
    RevGrad \cite{GaninL15}&75.0&86.0&96.2&87.0&74.3&91.5&85.0\\
    JAN \cite{JAN}&76.8&88.0&94.7&89.5&74.2&91.7&85.8\\
    MADA \cite{PeiCLW18}&75.0&87.9&96.0&88.8&75.2&92.2&85.8\\
    iCAN \cite{zhang2018collaborative}& \textbf{79.5}&89.7&94.7&89.9&\textbf{78.5}&92.0&87.4\\
    CDAN-RM \cite{long2017conditional}&77.2&88.3&\textbf{98.3}&90.7&76.7&\textbf{94.0}&87.5\\
    CDAN-M \cite{long2017conditional}&76.2&\textbf{89.5}&96.0&91.2&75.0&93.5&86.9\\
    \hline
   SCA &78.1&{89.2}&96.8&\textbf{91.3}&78.2&\textbf{94.0}&\textbf{87.9}\\
    \hline
  \end{tabular}
  \end{small}
  \end{center}
    \caption{Comparision of different methods for unsupervised domain adaptation on the ImageCLEF-DA dataset. The best results are in \textbf{bold}.}
      \label{image}
\end{table}

\textbf{Comparison on the Office-31 dataset.} We compare the proposed method with the recent state-of-the-art methods in Table \ref{office}.  Our method (SCA) gains 87.6\% accuracy, which is the second best performance on the Office-31 dataset.
Note that our method is comparable with CDAN-M \cite{long2017conditional} (87.6\% vs. 87.7\%). Besides, our method achieves the highest performance on three tasks (A $\rightarrow$ W, W $\rightarrow$ A, and W $\rightarrow$ D).
Our method is higher than MADA \cite{PeiCLW18} (87.6\% vs. 85.2\%).
Moreover, our method outperforms SimNet, iCAN, and JAN by 1.4\%, 0.4\%, and 3.2\%, respectively.

\textbf{Comparison on the ImageCLEF-DA dataset.} In Table \ref{image}, we compare the proposed method with state-of-the-art methods. SCA obtains 87.9\%, which outperforms the other methods. The accuracy of our method is 0.4\% higher than the second best method CDAN-RM \cite{long2017conditional}. 
Moreover, the proposed method respectively outperforms the MADA \cite{PeiCLW18},  iCAN \cite{zhang2018collaborative}, and  JAN \cite{JAN} by 2.1\%, 0.5\%, and 2.1\%. Specifically, our methods achieves the highest performance on two tasks (C $\rightarrow$ I and P $\rightarrow$ C).

The comparisons on the Office-31 dataset (Table \ref{office}) and the ImageCLEF-DA dataset (Table \ref{image}) demonstrate the effectiveness of the proposed method.

\setlength{\tabcolsep}{3.6pt}
\begin{table}[!t]
  \centering
  \label{table:office31}
  \begin{footnotesize}
  \begin{tabular}{l| c|c| c| c| c| c|c}
    \hline
   Method & A$\rightarrow$W & A$\rightarrow$ D& W$\rightarrow$A & W$\rightarrow$D & D$\rightarrow$A & D$\rightarrow$W & Avg. \\
    \hline
    \hline
   B (Basel.)&76.5&78.0&64.0&99.0&65.0&94.8&79.6 \\
   B + D& 87.2&84.9&69.8&99.2&67.8&96.5&84.2\\
   B + S& 85.0&87.0&67.2&99.4&67.5&98.2&84.1\\
   \hline
   SCA&93.5&89.5&72.7&100.0&72.4&97.5&87.6\\
    \hline
  \end{tabular}
  \end{footnotesize}
  \caption{Ablation experimental results of SCA.  The results are on the Office-31 dataset. ``B'' (Basel.) denotes the baseline trained only the source dataset, `` S'' represents the similarity-preserving constraint, and ``D'' denotes the domain-level alignment. SCA is the full system (``B + D + S'').}
    \label{office-ab}
\end{table}

\begin{figure}[!hpt]
\begin{center}
\includegraphics[width=0.9 \linewidth]{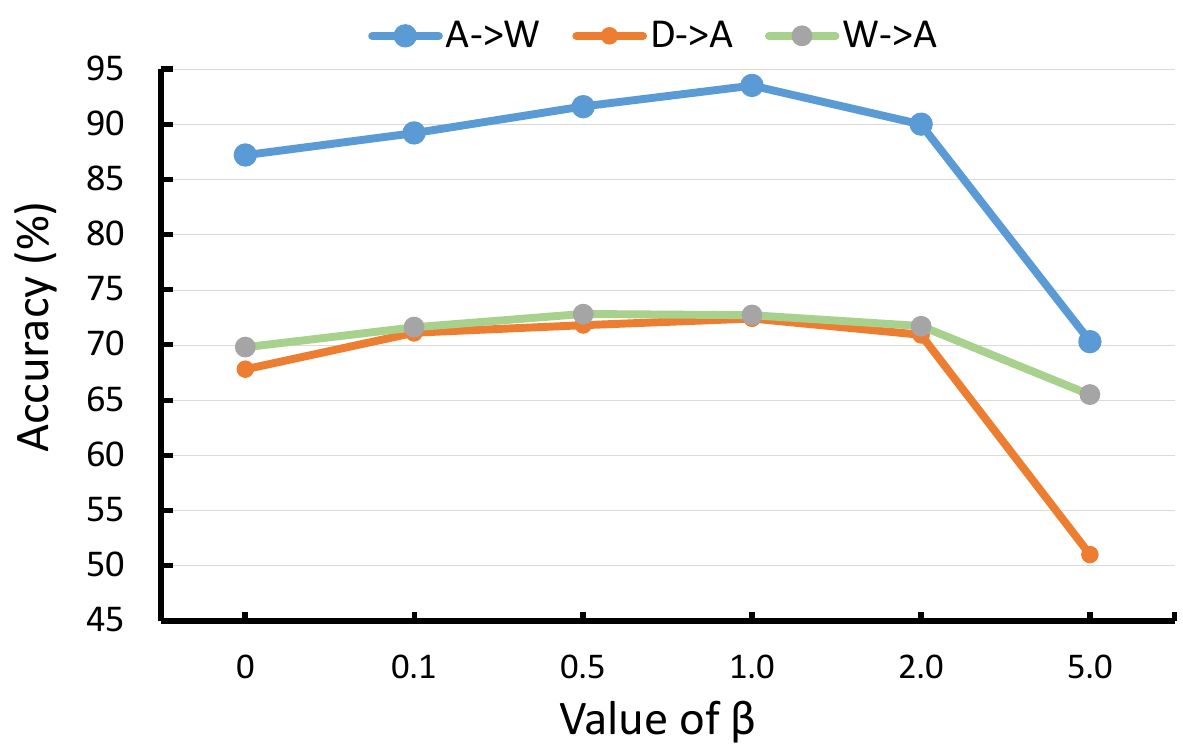}
\end{center}
\caption{Sensitivity to parameter $\beta$ (weight of the similarity-preserving constraint) in Eq. \ref{loss:all}. A larger $\beta$ means that the constraint has a greater impact on the distribution alignment. }
\label{fig:weight}
\end{figure}
\subsection{Component analysis}
In this section, we present step-by-step evaluation to analyze the effectiveness of SCA.

\textbf{Ablation study.} We investigate the impact of different components in SCA. We conduct the experiment on the Office-31 and report the results on Table \ref{office-ab}. 

The baseline is the network that we modify from ResNet-50, and it does not adopt any domain adaptation technique.
In this paper, we adopt JMMD \cite{JAN} for the domain-level alignment, and the result of ``B+ D'' is consistent with the experiment in \cite{JAN}.  Compared with ``B'' (Basel.),  ``B + D'' achieves higher performance, which indicates that it has ability to reduce the distribution discrepancy.  

On the top of domain-level alignment, the similarity-preserving constraint further brings +3.4\% improvement in average accuracy. This well demonstrates the importance of preserving underlying difference and commonness among source and target images.

As discussed in \ref{Discussion}, the similarity-preserving constraint can be viewed as a way to align distributions at class level. We further study its impact on the transfer accuracy, and report its results (``B+S'') in Table \ref{office-ab}. We can observe that only adopting the similarity constraint can also improve the baseline performance： it gains +4.5\% improvement over the baseline in average accuracy. This indicts that preserving class-level relations benefits the transfer accuracy. 

\begin{figure}[t]
\setlength{\belowcaptionskip}{0.2cm}
\begin{center}
\includegraphics[width=0.9 \linewidth]{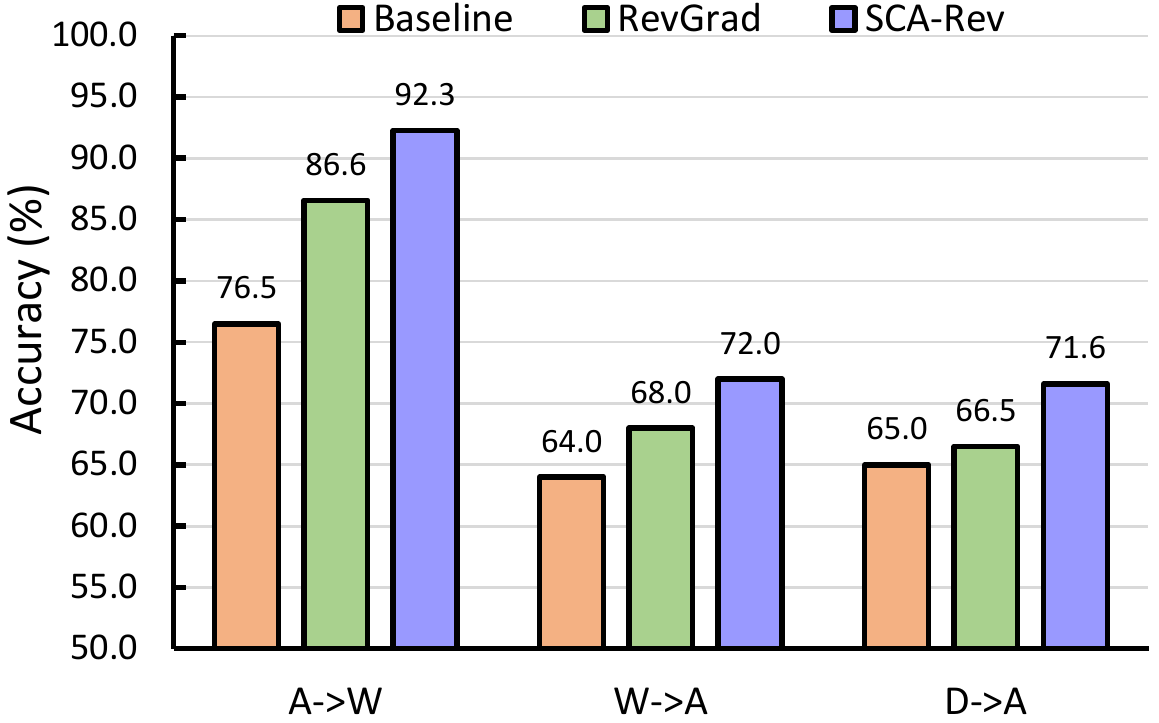}
\end{center}
\caption{Performance of various methods on three tasks (\textbf{A} $\rightarrow$ \textbf{W}, \textbf{W} $\rightarrow$ \textbf{A}, and \textbf{D} $\rightarrow$ \textbf{A}) of Office-31. Reverse Gradient (RevGrad) \cite{GaninL15} is a domain-level alignment method based on adversarial learning. SCA-Rev is the similarity-preserving alignment based on RevGrad.}
\label{fig:method}
\end{figure}
\textbf{Weight of the similarity-preserving constraint.} The $\beta $ in Eq. \ref{loss:all} control the importance of similarity-preserving constrain. A larger $\beta$ means that the constraint has a greater impact on the distribution alignment. In Fig \ref{fig:weight}, we demonstrate the transfer accuracy of SCA by varying the $\beta \in \{0, 0.1, 0.5, 1, 2, 5\}$ on three tasks, \textbf{A} $\to$ \textbf{W},  \textbf{W} $\to$ \textbf{A}, and  \textbf{D} $\to$ \textbf{A}. Note the when $\beta$ is set to 0, the similarity-preserving constrain has no impact. As shown in Fig. \ref{fig:weight}, when the $\beta$ increases from 0 to 1, the performance on three tasks grow and reach the best at $\beta=1$. However, when the $\beta$ is too large ($\beta$=5), the accuracy will drop by a large margin. Empirically, the best parameter $\beta$ is between 0.5 to 2 in our method.

\textbf{Domain-level alignment method.} As discussed in Section \ref{ga}, we use a discrepancy-based metric JMMD for domain-level alignment.
We note that the proposed similarity-preserving constraint can work collaboratively with other domain-level alignment methods.  To validate this, we conduct the experiment on three tasks of Office-31: \textbf{A} $\rightarrow$ \textbf{W}, \textbf{W} $\rightarrow$ \textbf{A}, and \textbf{D} $\rightarrow$ \textbf{A}. 
We adopt an adversarial adaptation method named Reverse Gradient (RevGrad) \cite{GaninL15} for domain-level alignment. Based on RevGrad, we construct the similarity constrained alignment network (SCA-Rev), and report the results on the Fig. \ref{fig:method}.

 As shown in Fig. \ref{fig:method}, RevGrad can improve the accuracy of baseline, which indicates it has ability to reduce the distribution discrepancy. Moreover, SCA-Rev further improves the accuracy of RevGrad. SCA-Rev gains +5.7\%, +4.0\% and 5.1\% improvements over RevGrad on \textbf{A} $\rightarrow$ \textbf{W}, \textbf{W} $\rightarrow$ \textbf{A}, and \textbf{D} $\rightarrow$ \textbf{A}, respectively.
On the one hand, the results demonstrate that preserving the two class-level relations is crucial for the domain-level alignment. On the other hand, these results indicate that the similarity-preserving constraint can work collaboratively with other domain-level alignment methods.

\begin{figure}[t]
\begin{center}
\includegraphics[width=0.8 \linewidth]{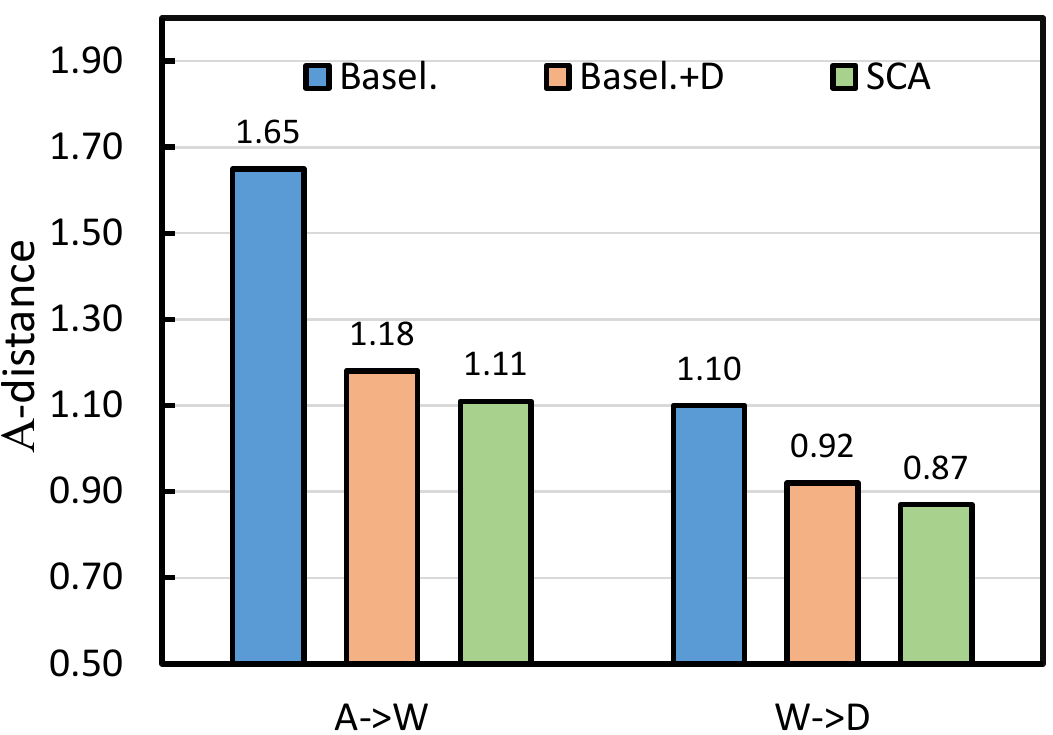}
\end{center}
\caption{Distribution discrepancy measured by $\mathcal{A}$-distance on tasks \textbf{A} $\rightarrow$ \textbf{W} and \textbf{W} $\rightarrow$ \textbf{D}. Three methods are compared: (a) baseline (Basel.), (b) domain-level alignment  (Basel. + D), and (c) SCA.}
\label{fig:adistance}
\end{figure}

\textbf{Distribution discrepancy.}
The domain adaptation theory \cite{DavidBCKPV10, MansourMR09} introduces $\mathcal{A}$-distance to measure the distribution discrepancy . The ${\cal{A}}$-distance is defined as ${d_{\cal A}} = 2\left( {1 - 2\epsilon } \right)$, where $\epsilon$ is the generalization error of a classifier trained to discriminate source and target. 
We report the ${d_{\cal A}}$ on two tasks (\textbf{A} $\rightarrow$ \textbf{W}, \textbf{W} $\rightarrow$ \textbf{D}) of Office-31 with features of baseline, domain-level alignment (basel. + G), and SCA. 
As shown in Fig. \ref{fig:adistance}, ${d_{\cal A}}$ on SCA features is much smaller than ${d_{\cal A}}$ on the baseline and domain-level alignment features. This indicates that SCA features can reduce the distribution discrepancy more effectively.

\section{Conclusion and Future Work}
In this paper, we present the similarity constrained alignment (SCA) method to address the semantic misalignment problem.
SCA enforces a similarity-preserving constraint to maintain the underlying difference and commonness among the source and target images. In the absence of target labels, we use a classifier trained on source images to assign pseudo labels to the target images. Given labeled source images and pseudo-labeled target images, the similarity-preserving constraint can be implemented by minimizing the triplet loss.
Under the collaborative supervision of the domain alignment loss and the triplet loss, SCA learns domain-invariant embeddings with two important properties, \ie, intra-class compactness and inter-class separability. Thus, the distributions can be aligned at both domain and class level, which alleviates the semantic misalignment problem.
The experimental results on two benchmarks demonstrate that the proposed SCA is effective and competitive with the state-of-the-art methods.
In the future, we will extend this idea to multiple target domains, where the class-level relations among multi-domains will be explored. 

{\small
\bibliographystyle{ieee}
\bibliography{egbib}

\begin{thebibliography}{10}\itemsep=-1pt

\bibitem{DavidBCKPV10}
S.~Ben{-}David, J.~Blitzer, K.~Crammer, A.~Kulesza, F.~Pereira, and J.~W.
  Vaughan.
\newblock A theory of learning from different domains.
\newblock {\em Machine Learning}, 79(1-2):151--175, 2010.

\bibitem{DBLP:journals/corr/BousmalisSDEK16}
K.~Bousmalis, N.~Silberman, D.~Dohan, D.~Erhan, and D.~Krishnan.
\newblock Unsupervised pixel-level domain adaptation with generative
  adversarial networks.
\newblock In {\em {IEEE} Conference on Computer Vision and Pattern
  Recognition}, 2017.

\bibitem{CaoMLW18}
Z.~Cao, L.~Ma, M.~Long, and J.~Wang.
\newblock Partial adversarial domain adaptation.
\newblock In {\em ECCV}, 2018.

\bibitem{ChenWB11}
M.~Chen, K.~Q. Weinberger, and J.~Blitzer.
\newblock Co-training for domain adaptation.
\newblock In {\em NIPS}, 2011.

\bibitem{ChenSG13}
X.~Chen, A.~Shrivastava, and A.~Gupta.
\newblock {NEIL:} extracting visual knowledge from web data.
\newblock In {\em ICCV}, 2013.

\bibitem{ChopraHL05}
S.~Chopra, R.~Hadsell, and Y.~LeCun.
\newblock Learning a similarity metric discriminatively, with application to
  face verification.
\newblock In {\em CVPR}, 2005.

\bibitem{image-image18}
W.~Deng, L.~Zheng, Q.~Ye, G.~Kang, Y.~Yang, and J.~Jiao.
\newblock Image-image domain adaptation with preserved self-similarity and
  domain-dissimilarity for person re-identification.
\newblock In {\em {IEEE} Conference on Computer Vision and Pattern
  Recognition}, 2018.

\bibitem{FernandoHST13}
B.~Fernando, A.~Habrard, M.~Sebban, and T.~Tuytelaars.
\newblock Unsupervised visual domain adaptation using subspace alignment.
\newblock In {\em ICCV}, 2013.

\bibitem{GaninL15}
Y.~Ganin and V.~S. Lempitsky.
\newblock Unsupervised domain adaptation by backpropagation.
\newblock In {\em ICML}, 2015.

\bibitem{GaninUAGLLML16}
Y.~Ganin, E.~Ustinova, H.~Ajakan, P.~Germain, H.~Larochelle, F.~Laviolette,
  M.~Marchand, and V.~S. Lempitsky.
\newblock Domain-adversarial training of neural networks.
\newblock {\em Journal of Machine Learning Research}, 17:59:1--59:35, 2016.

\bibitem{GoldbergerRHS04}
J.~Goldberger, S.~T. Roweis, G.~E. Hinton, and R.~Salakhutdinov.
\newblock Neighbourhood components analysis.
\newblock In {\em NIPS}, 2004.

\bibitem{GoodfellowPMXWOCB14}
I.~J. Goodfellow, J.~Pouget{-}Abadie, M.~Mirza, B.~Xu, D.~Warde{-}Farley,
  S.~Ozair, A.~C. Courville, and Y.~Bengio.
\newblock Generative adversarial networks.
\newblock In {\em NIPS}, 2014.

\bibitem{GopalanLC11}
R.~Gopalan, R.~Li, and R.~Chellappa.
\newblock Domain adaptation for object recognition: An unsupervised approach.
\newblock In {\em ICCV}, 2011.

\bibitem{GrettonBRSS06}
A.~Gretton, K.~M. Borgwardt, M.~J. Rasch, B.~Sch{\"{o}}lkopf, and A.~J. Smola.
\newblock A kernel method for the two-sample-problem.
\newblock In {\em NIPS}, 2006.

\bibitem{resnet}
K.~He, X.~Zhang, S.~Ren, and J.~Sun.
\newblock Deep residual learning for image recognition.
\newblock In {\em CVPR}, 2016.

\bibitem{HermansBL17}
A.~Hermans, L.~Beyer, and B.~Leibe.
\newblock In defense of the triplet loss for person re-identification.
\newblock {\em CoRR}, abs/1703.07737, 2017.

\bibitem{hoffman2018cycada}
J.~Hoffman, E.~Tzeng, T.~Park, J.-Y. Zhu, P.~Isola, K.~Saenko, A.~A. Efros, and
  T.~Darrell.
\newblock Cycada: Cycle-consistent adversarial domain adaptation.
\newblock In {\em {International} Conference on Machine Learning}, 2018.

\bibitem{HuLT14}
J.~Hu, J.~Lu, and Y.~Tan.
\newblock Discriminative deep metric learning for face verification in the
  wild.
\newblock In {\em CVPR}, 2014.

\bibitem{kang2018deep}
G.~Kang, L.~Zheng, Y.~Yan, and Y.~Yang.
\newblock Deep adversarial attention alignment for unsupervised domain
  adaptation: the benefit of target expectation maximization.
\newblock In {\em ECCV}, 2018.

\bibitem{KulisSD11}
B.~Kulis, K.~Saenko, and T.~Darrell.
\newblock What you saw is not what you get: Domain adaptation using asymmetric
  kernel transforms.
\newblock In {\em CVPR}, 2011.

\bibitem{LaineA16}
S.~Laine and T.~Aila.
\newblock Temporal ensembling for semi-supervised learning.
\newblock {\em CoRR}, abs/1610.02242, 2016.

\bibitem{LiF10}
L.~Li and F.~Li.
\newblock {OPTIMOL:} automatic online picture collection via incremental model
  learning.
\newblock {\em International Journal of Computer Vision}, 2010.

\bibitem{DBLP:conf/nips/LiuT16}
M.~Liu and O.~Tuzel.
\newblock Coupled generative adversarial networks.
\newblock In {\em {Advances} in Neural Information Processing Systems}, 2016.

\bibitem{LiuWYLRS17}
W.~Liu, Y.~Wen, Z.~Yu, M.~Li, B.~Raj, and L.~Song.
\newblock Sphereface: Deep hypersphere embedding for face recognition.
\newblock In {\em CVPR}, 2017.

\bibitem{DAN}
M.~Long, Y.~Cao, J.~Wang, and M.~I. Jordan.
\newblock Learning transferable features with deep adaptation networks.
\newblock In {\em ICML}, 2015.

\bibitem{long2017conditional}
M.~Long, Z.~Cao, J.~Wang, and M.~I. Jordan.
\newblock Conditional adversarial domain adaptation.
\newblock In {\em NIPS}, 2018.

\bibitem{RTN}
M.~Long, H.~Zhu, J.~Wang, and M.~I. Jordan.
\newblock Unsupervised domain adaptation with residual transfer networks.
\newblock In {\em NIPS}, 2016.

\bibitem{JAN}
M.~Long, H.~Zhu, J.~Wang, and M.~I. Jordan.
\newblock Deep transfer learning with joint adaptation networks.
\newblock In {\em ICML}, 2017.

\bibitem{MansourMR09}
Y.~Mansour, M.~Mohri, and A.~Rostamizadeh.
\newblock Domain adaptation: Learning bounds and algorithms.
\newblock In {\em {COLT} 2009 - The 22nd Conference on Learning Theory,
  Montreal, Quebec, Canada, June 18-21, 2009}, 2009.

\bibitem{MotiianPAD17}
S.~Motiian, M.~Piccirilli, D.~A. Adjeroh, and G.~Doretto.
\newblock Unified deep supervised domain adaptation and generalization.
\newblock In {\em ICCV}, 2017.

\bibitem{PeiCLW18}
Z.~Pei, Z.~Cao, M.~Long, and J.~Wang.
\newblock Multi-adversarial domain adaptation.
\newblock In {\em AAAI}, 2018.

\bibitem{simnet}
P.~O. Pinheiro.
\newblock Unsupervised domain adaptation with similarity learning.
\newblock In {\em CVPR}, 2018.

\bibitem{distill}
I.~Radosavovic, P.~Doll{\'{a}}r, R.~B. Girshick, G.~Gkioxari, and K.~He.
\newblock Data distillation: Towards omni-supervised learning.
\newblock In {\em CVPR}, 2018.

\bibitem{ILSVRC15}
O.~Russakovsky, J.~Deng, H.~Su, J.~Krause, S.~Satheesh, S.~Ma, Z.~Huang,
  A.~Karpathy, A.~Khosla, M.~Bernstein, A.~C. Berg, and L.~Fei-Fei.
\newblock {ImageNet Large Scale Visual Recognition Challenge}.
\newblock {\em International Journal of Computer Vision (IJCV)},
  115(3):211--252, 2015.

\bibitem{SaenkoKFD10}
K.~Saenko, B.~Kulis, M.~Fritz, and T.~Darrell.
\newblock Adapting visual category models to new domains.
\newblock In {\em ECCV}, 2010.

\bibitem{SaitoUH17}
K.~Saito, Y.~Ushiku, and T.~Harada.
\newblock Asymmetric tri-training for unsupervised domain adaptation.
\newblock In {\em ICML}, 2017.

\bibitem{SchroffKP15}
F.~Schroff, D.~Kalenichenko, and J.~Philbin.
\newblock Facenet: {A} unified embedding for face recognition and clustering.
\newblock In {\em CVPR}, 2015.

\bibitem{SongXJS16}
H.~O. Song, Y.~Xiang, S.~Jegelka, and S.~Savarese.
\newblock Deep metric learning via lifted structured feature embedding.
\newblock In {\em CVPR}, 2016.

\bibitem{DBLP:conf/aaai/SunFS16}
B.~Sun, J.~Feng, and K.~Saenko.
\newblock Return of frustratingly easy domain adaptation.
\newblock In {\em {Proceedings} of AAAI Conference on Artificial Intelligence},
  2016.

\bibitem{SunS16}
B.~Sun and K.~Saenko.
\newblock Deep {CORAL:} correlation alignment for deep domain adaptation.
\newblock In {\em ECCV Workshops}, 2016.

\bibitem{DBLP:conf/cvpr/TorralbaE11}
A.~Torralba and A.~A. Efros.
\newblock Unbiased look at dataset bias.
\newblock In {\em {IEEE} Conference on Computer Vision and Pattern
  Recognition}, 2011.

\bibitem{TzengHSD17}
E.~Tzeng, J.~Hoffman, K.~Saenko, and T.~Darrell.
\newblock Adversarial discriminative domain adaptation.
\newblock In {\em CVPR}, 2017.

\bibitem{TzengHZSD14}
E.~Tzeng, J.~Hoffman, N.~Zhang, K.~Saenko, and T.~Darrell.
\newblock Deep domain confusion: Maximizing for domain invariance.
\newblock {\em CoRR}, abs/1412.3474, 2014.

\bibitem{WeinbergerS09}
K.~Q. Weinberger and L.~K. Saul.
\newblock Distance metric learning for large margin nearest neighbor
  classification.
\newblock {\em Journal of Machine Learning Research}, 10, 2009.

\bibitem{WenZL016}
Y.~Wen, K.~Zhang, Z.~Li, and Y.~Qiao.
\newblock A discriminative feature learning approach for deep face recognition.
\newblock In {\em ECCV}, 2016.

\bibitem{zhang2018collaborative}
W.~Zhang, W.~Ouyang, W.~Li, and D.~Xu.
\newblock Collaborative and adversarial network for unsupervised domain
  adaptation.
\newblock In {\em CVPR}, 2018.

\end{thebibliography}
}

\end{document}